\documentclass[conference]{IEEEtran}
\IEEEoverridecommandlockouts
\usepackage[pagebackref=true,breaklinks=true,colorlinks,bookmarks=false]{hyperref}

\usepackage{times}
\usepackage{epsfig}
\usepackage{latexsym}
\usepackage{graphicx}
\usepackage{url}
\usepackage{amsmath}
\usepackage{amssymb,soul}
\usepackage{multirow}
\usepackage{xcolor}
\usepackage{enumitem}
\usepackage{xparse}
\usepackage{booktabs}
\usepackage{etoolbox}
\usepackage{cleveref}

\newcommand{\ctab}[1]{\begin{tabular}{c} #1 \end{tabular}}
\newcommand{\etal}{\textit{et al.}}

\usepackage{arydshln}               
\definecolor{RoseQuartzBg}{HTML}{F7CAC9}
\definecolor{RoseQuartz}{HTML}{F5A798}
\definecolor{Serenity}{HTML}{92A8D1}
\definecolor{themered}{HTML}{FF8375}

\NewDocumentCommand{\heng}{ mO{} }{\textcolor{RoseQuartz}{\textsuperscript{\textit{Heng}}\textsf{\textbf{\small[#1]}}}}
\NewDocumentCommand{\pengfei}{ mO{} }{\textcolor{Serenity}{\textsuperscript{\textit{Pengfei}}\textsf{\textbf{\small[#1]}}}}
\NewDocumentCommand{\Xiaodan}{ mO{} }{\textcolor{Serenity}{\textsuperscript{\textit{Xiaodan}}\textsf{\textbf{\small[#1]}}}}
\NewDocumentCommand{\spencer}{ mO{} }{\textcolor{purple}{\textsuperscript{\textit{Spencer}}\textsf{\textbf{\small[#1]}}}}
\NewDocumentCommand{\shi}{ mO{} }{\textcolor{RoseQuartz}{\textsuperscript{\textit{Shi}}\textsf{\textbf{\small[#1]}}}}

\NewDocumentCommand{\method}{ mO{} }{#1}

\newcommand{\NumPortraits}[1]{\ifstrequal{#1}{all}{51,939}{\ifstrequal{#1}{wikiart}{25,588}{\ifstrequal{#1}{wikidata}{26,351}{Undefined}}}}
\newcommand{\Model}{MUSE}
\newcommand{\NumAttributes}{11}

\date{}

\begin{document}

\title{MUSE: Textual Attributes Guided Portrait Painting Generation}





 \author{Xiaodan Hu\textsuperscript{\textnormal{1}}, Pengfei Yu\textsuperscript{\textnormal{1}}, Kevin Knight\textsuperscript{\textnormal{2}}, Heng Ji\textsuperscript{\textnormal{1}}, Bo Li\textsuperscript{\textnormal{1}}, Honghui Shi\textsuperscript{\textnormal{1}}\\
  \textsuperscript{1}University of Illinois at Urbana-Champaign  \\
  \textsuperscript{2}DiDi Labs\\
  \texttt{\fontfamily{pcr}\selectfont\{xiaodan8,hengji\}@illinois.edu},\\ \texttt{\fontfamily{pcr}\selectfont{kevinknight@didiglobal.com}}
}

\maketitle
\begin{abstract}

We propose a novel approach, \emph{\Model{}}, to automatically generate portrait paintings guided by textual attributes. 
\emph{\Model{}} takes a set of attributes written in text, in addition to facial features extracted from a photo of the subject as input. 
We propose \NumAttributes{} 
attribute types to represent inspirations from a subject's profile, emotion, story, and environment. 
Then we design a novel stacked neural network architecture by extending an image-to-image generative model to accept textual attributes. 
Experiments show that our approach significantly outperforms several state-of-the-art methods without using textual attributes, 
with Inception Score score increased by 6\% and Fr\'echet Inception Distance (FID) score decreased by 11\%, respectively.
We also propose a 
new attribute reconstruction metric to evaluate whether the generated portraits preserve the subject's attributes. Experiments show that our approach can accurately illustrate 78\% textual attributes, which also 
help \emph{\Model{}} capture the subject in a more creative and expressive way. 
\footnote{We have made all of the data sets, resources and programs related to this new benchmark available at \url{https://github.com/xiaodanhu/MUSE}.}

\end{abstract}

\section{Introduction}

\begin{table*}[!ht]
\small
\centering
\begin{tabular}{lll}
	\toprule
	\textbf{Text Attribute} & \textbf{Value Examples} & \textbf{\#Values} \\
	\midrule
    Age & Child; Young adults; Middle-aged adults; Older adults & 4\\
	Clothing & Blazer; Coat and Jacket; Choir \& Religious Robe; Dressing Gown; Dress; Shirt; Sweater & 14\\
	Facial Expression & Smile; Smirk; Sneer; Glance; Wink; Wrinkle the nose; Long face; Blank expression & 14\\
	Gender & Male; Female; Other & 3\\
	Hair & Straight hair; Wavy hair; Black hair; Blond hair; Short hair & 5\\
    Mood & Calm; Excited; Happy; Angry; Apathetic; Sad & 6\\
	Pose \& Gesture & Lying; Sitting; Squatting or crouching; Standing; Riding; Shooting; Sleeping; Bowing & 9\\
	Setting & In the room; In the hallway; On the street; At the river; In the courtyard or a garden & 9\\
	Style & Impressionism; Realism; Classical Art; Modernism; Chinese paintings; Japanese paintings & 7 \\
	Time & Before 1970; After 1970 & 2 \\
	Weather & Rainy; Stormy; Sunny; Cloudy; Hot; Cold; Windy; Foggy; Snow & 9\\
	\bottomrule
\end{tabular}
\caption{Text Attributes to Represent Portrait Inspirations.
}
\vspace{-7pt}
\label{tab:dimensions}
\end{table*}

\begin{figure*}
\centering
\begin{tabular}{c}
\includegraphics[height=6.3cm,keepaspectratio]{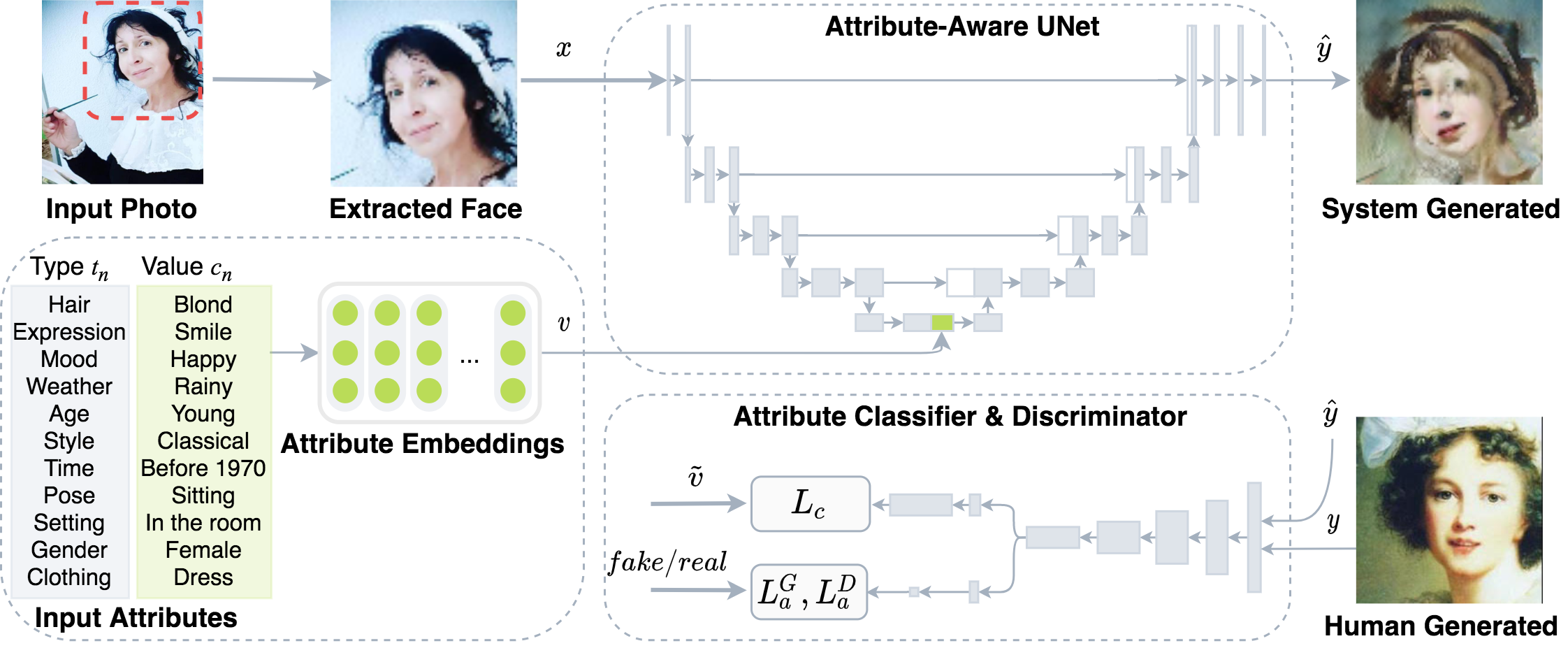}
\end{tabular}
\caption{Overview of the model architecture. Given an input photo $x$ and attribute embeddings $\mathbf{v}$, the model can generate the portrait $\hat{y}$. The generator $G$ is an attribute-aware UNet that incorporate attribute embeddings in hidden representation. The discriminator $D$ is a stack of convolutional layers followed by fully connected layers. By using the adversarial loss $L_a$ and the attribute classification loss $L_c$, the discriminator $D$ can both recognize the realness of the generated portrait $\hat{y}$ and classify $\hat{y}$ into different class of each attribute. 
}
\vspace{-15pt}
\label{fig:overview}
\end{figure*}





We aim to teach computer to automatically generate portrait paintings, guided by textual attributes. Portrait is a special genre in painting, where the goal is to present not only the outward appearance of a specific human subject, but also their inner significance inspired by admiration or affection for the subject. A good portrait needs to be realistic, and thus it's important to take a photo of the subject as input. Some recent attempts \cite{Liao2017ACM,Zhu2017CycleGAN,Zhu2017BicycleGAN,Lee2018DRIT} try to automate the conversion from photo to portrait. Most of these methods are only based on visual style transfer
~\cite{Liao2017ACM,Zhu2017CycleGAN,Ancient2019}. 
However, art works reflect not only the artist's hard work and dexterous technique but also often carry their personal emotions and memories on the subjects, due to the intimate relationship between the artist and the subject, either bound together before or during the painting. 
As Aristotle stated, ``The aim of Art is to present not the outward appearance of things, but their inner significance; for this, not the external manner and detail, constitutes true reality.''~\cite{Aymar1967}. 
From a good portrait, we can often reveal the story of the subject's life, such as hobby, personality, mood, or a special occasion which may even involve the artist, from a certain facial expression, hairstyle, 
or the artist's clever use of colors and lines.


We represent a story (i.e. the inspirations) of the subject with 11 text attributes, as shown in Table~\ref{tab:dimensions}. 
We design a new portrait generation framework called \emph{\Model{}}, which takes these inspirational textual attributes in addition to face regions as input for portrait generation. 
We first feed the extracted face from the input photo into the encoder part of a UNet \cite{UNet2015}, a convolutional network architecture that 
has show promising results on 
image generation~\cite{AttGAN2019,UNet2019}. Then we directly feed the textual attributes into an attribute encoder. 
Finally the textual attribute embeddings are integrated with the hidden representation of the input photo 
as an input to the attribute-based decoder to generate the portrait. Figure \ref{fig:overview} illustrates an example where the facial expression and hair style are automatically changed based on a set of textual attributes.

Moreover, existing evaluation metrics such as Inception Score (IS)~\cite{Incept_score} only check the overlapped visual content between the machine-generated and human-generated portraits. But this is not how human assessors approach and appreciate a painting. 
We design a novel metric 
to evaluate how many text attributes human assessors can reconstruct from the system generated portraits. 
Experiments show that our method outperforms state-of-the-art on all measures, and portrait generation is an effective way to acquire and illustrate textual attributes.  
In summary, the main contributions of this paper are as follows: 
\begin{itemize}
\item We propose the first inspiration-to-portrait generation framework \emph{\Model{}} that takes text description of attributes into account to generate portraits aligned with its background story such as underlying emotions. 
\item 
We develop a novel neural network architecture incorporating textual attributes for portrait generation. 
Rather than using a binary sequence of facial attributes as input, we apply attribute embeddings, which are initialized from portrait data and optimized during training. 
Instead of preserving the attributes of input photos, we design a novel discriminator to encourage diversity and realism.
\item We create a large portrait generation data set containing 3,928 photo-portrait pairs with manually annotated attributes as a new benchmark, along with \NumPortraits{all} portraits without annotations, and will share the resources with the community for further exploration.
\item We propose a novel evaluation metric based on human attribute 
reconstruction 
to better assess the quality of generated portraits.
\end{itemize}
\section{MUSE: Portrait Generation Approach}


\emph{\Model{}} takes two sources of input to generate portraits: (1) a photo $x$ of the subject; (2) a set of textual attributes in form of type-value pairs $\{(t_i,c_i)\}_{i=1}^n$. 
\emph{\Model{}} contains a 
generator $G$ and a corresponding discriminator $D$ for adversarial training. 
To demonstrate the importance and effectiveness of textual attributes, we modify a state-of-the-art image generator, UNet (\cite{UNet2015,UNet2019}),  
to incorporate attribute embeddings $v$ into generation. The discriminator $D$ takes the system generated portrait $\hat{y}$ together with its corresponding human generated portrait $y$ to evaluate the generator $G$. The overall architecture of \emph{\Model{}} is depicted in Figure \ref{fig:overview}.  

\vspace{-5pt}
\subsection{Attribute Embedding}
\label{sec:EntityTyping}


We propose 11 textual attributes to represent the inspirations, 
as shown in Table~\ref{tab:dimensions}. We select these attributes from various 
online resources including 
the LitCharts Library, ClarkandMiller.com, and ManyThings.org.

We assign each type-value pair $(t_i, c_i)$ a unique embedding $\mathbf{v}_i$. Given a set of $n$ attribute type-value pairs $\{(t_i, c_i)\}_{i=1}^n$ as input, we concatenate embeddings for $n$ pairs to get the input attribute embedding $\mathbf{v}=[\mathbf{v}_1;\mathbf{v}_2;\ldots;\mathbf{v}_n]\in\mathbb{R}^{nd_w}$. 

The attribute values in Table~\ref{tab:dimensions} are correlated. For instance, 
the clothing under 
\texttt{rainy} weather is more likely to be \texttt{blazer} or \texttt{coat}. To capture such inter-dependency between multiple attribute values, we initialize 
$\mathbf{v}_i$ by using
domain-specific attribute embeddings trained from portrait data, where we use skip-gram methods in Word2Vec \cite{mikolov2013efficient} 
and consider attribute values corresponding to the same portrait as a bag of words in one context window.

\vspace{-10pt}
\subsection{Attribute-aware UNet}
\label{sec:ImageEncoder}

UNet (\cite{UNet2015,UNet2019}) 
is a high-quality image-to-image generative model.
Given an input photo $x$, UNet first encodes $x$ into a hidden representation $\mathbf{h}\in\mathbb{R}^{d_h}=G_{enc}(x)$ using multi-layer Convolutional Neural Networks (CNNs), which is further decoded into an output image $\hat{y}=G_{dec}(\mathbf{h})$ using a stack of transposed convolutional layers.

Our generator $G$ employs the UNet architecture. The input photo $x$ is first encoded into the hidden representation $\mathbf{h}$. Then 
we aggregate the hidden representation $\mathbf{h}$ 
and attribute embeddings $\mathbf{v}$ of the expected portrait as
\begin{equation*}
    \mathbf{h}^a = \sigma \left(\mathbf{W}_h\mathbf{h} + \mathbf{W}_v\mathbf{v}+\mathbf{b}\right),
\end{equation*}
where $\mathbf{W}_h\in\mathbb{R}^{d_h\times d_h}$, $\mathbf{W}_v\in\mathbb{R}^{d_h\times nd_w}$ and $\mathbf{b}\in\mathbb{R}^{d_h}$ are learnable parameters. $\sigma$ is an activation function and we use $\mathrm{ReLU}$ \cite{relu2010}. 
We decode the portrait as $\hat{y}=G_{dec}(\mathbf{h}^a)$. For simplicity, we use $G_{enc}^a(x,\mathbf{v})$ to represent $\mathbf{h}^a$.

\subsection{Loss Functions}
\label{sec:LossFunction}




We apply adversarial training simultaneously for the generator and discriminator $(G, D)$ to learn the mapping from the input photos $X$ and input attribute embeddings $V$ to the output portraits $Y$. Given training samples $\{(x^{(i)},\mathbf{v}^{(i)}, y^{(i)})\}^N_{i=1}$ where the input photo $x^{(i)} \in X$, input attribute embeddings $\mathbf{v}^{(i)} \in V$ and the human generated portrait $y^{(i)} \in Y$, we denote the data distribution as $x \sim p_{\mathrm{data}(x)}$, $\mathbf{v} \sim p_{\mathrm{
attr}}$ and $y \sim p_{\mathrm{data}(y)}$. While $G$ tries to generate realistic portraits $\hat{y}$ similar to the portraits $y$ in $Y$ domain, $D$ tries to distinguish between $\hat{y}$ and $y$.

We compute the adversarial loss following GAN \cite{NIPSGAN}, with input photo $x$, attribute embeddings $\mathbf{v}$, and $y$ as corresponding human generated portraits rescaled to the same size as the outputs of $G$. Specifically, the adversarial losses of the generator $G_i$ and discriminator $D$ are 
as follows:
\begin{equation}
\mathcal{L}_a^G=-\mathbb{E}_{x \sim p_{\mathrm{data}(x)}, \mathbf{v} \sim p_{\mathrm{attr}}}\log D\left( G_{dec}(G_{enc}^a(x,\mathbf{v})) \right),
\label{eq:G}
\end{equation}

\begin{equation}
\begin{aligned}
\mathcal{L}_a^D &= \mathbb{E}_{x \sim p_{\mathrm{data}(x)}, \mathbf{v} \sim p_{\mathrm{attr}}}\log D\left( G_{dec}(G_{enc}^a(x,\mathbf{v})) \right) \\
&\quad - \mathbb{E}_{y \sim p_{\mathrm{data}}(y)} \log D(y).
\end{aligned}
\label{eq:D}
\end{equation}

In addition to the adversarial loss $\mathcal{L}_a^G$, 
we further use L1 distance to force the generator not only to generate realistic portraits to fool the discriminator but also get close to the human generated portrait. The L1 loss can be obtained as
\begin{equation}
\mathcal{L}_{L1} = \mathbb{E}_{x,\mathbf{v},y}{\left\Vert y- G_{dec}(G_{enc}^a(x,\mathbf{v}))\right\Vert_1},
\label{eq:L1}
\end{equation}

While the adversarial learning is employed on the system generated portrait $\hat{y}$ to ensure its visual reality, $\hat{y}$ is also expected to correctly contain the desired attributes $\mathbf{v}$. Hence, an attribute classifier $F$ is used to constrain the system generated portrait $\hat{y}$ with $\mathbf{v}$. Let $\mathbf{\tilde{v}}$ denote the one-hot vectors of $\mathbf{v}$, the attribute classification loss $L_c$ can be obtained as
\begin{equation}
\mathcal{L}_{c} = \mathbb{E}_{x \sim p_{\mathrm{data}(x)}, \mathbf{v} \sim p_{\mathrm{attr}}} [\rho (F\left( G_{dec}(G_{enc}^a(x, \mathbf{v})) \right), \mathbf{\tilde{v}})],
\label{eq:classification}
\end{equation}
where $\rho$ is the summation of binary cross-entropy losses of all attributes as follows: 
\begin{equation}
\rho(\mathbf{\hat{v}}, \mathbf{\tilde{v}}) = \sum^n_{i=1} -\mathbf{\tilde{v}}_i \log{\mathbf{\hat{v}}_i} -(1-\mathbf{\tilde{v}}_i)\log(1-\mathbf{\hat{v}}_i).
\label{eq:rho}
\end{equation}
where $\mathbf{\hat{v}}_i=F_i\left( G_{dec}(G_{enc}^a(x,\mathbf{v})) \right)$ indicates the prediction of the $i^{th}$ attribute $\mathbf{v}_i$.

Overall, by combining the adversarial loss and the attribute classification loss, the final objective functions of the generator $G$ and the discriminator $D$ are as follows:
\begin{equation}
\mathcal{L}^G = \mathcal{L}_a^G + \lambda_1 \mathcal{L}_{c} + \lambda_2 \mathcal{L}_{L1}.
\label{eq:loss_G}
\end{equation}
\vspace{-9pt}
\begin{equation}
\mathcal{L}^D = \mathcal{L}_{a}^D + \lambda_3 \mathcal{L}_{c}.
\label{eq:loss_D}
\end{equation}
where $\lambda_1,\lambda_2,\lambda_3$ are hyper-parameters that control the relative importance of the losses. 

\subsection{Training and Testing Process}
\label{sec:train_test}

In the training phase, we use photo-attribute-portrait training samples $\{(x^{(i)},\mathbf{v}^{(i)}, y^{(i)})\}^N_{i=1}$ to train our model. The generator $G$ takes $x$ and $\mathbf{v}$ as input and generate the portraits $\hat{y}$. The discriminator $D$ learns to
distinguish real and generated samples by taking the generated portraits from $G$ and human generated portraits $y$, respectively. 
In the testing phase, the photo-attribute pairs are fed into $G$ to generate portraits $\hat{y}$.
\section{Experiments}
\subsection{Data and Experiment Setting}
\label{sec:PortraitsDataset}

\noindent\textbf{Datasets.} 
We have collected 4,608 photo-portrait pairs from various sources including:  (1) the museum artwork remake challenges that requires people to re-create artworks at home (e.g., the Getty Museum Challenge \cite{Getty}
, Metropolitan Museum \cite{mettwinning}
, Pinchuk Art Centre \cite{pinchuk}
, and Rijksmuseum \cite{rijksmuseum}
; (2) Tussen Kunst \& Quarantaine Instagram \cite{TKQ}
that shares homemade recreations, (3) a remake project \cite{booooooom}
built by Booooooom \& Adobe to remake a famous work of art using photography; and (4) the remake images collected by Pinterest \cite{pinterest}. 
We manually remove non-human and bad quality photo-portrait pairs, and 
remove duplicated portraits. 
We use the Amazon Sandbox platform to perform dual attribute annotations and careful adjudication on the portraits. 
3,296 high resolution pairs are selected and cropped to include only the face region. We use 3,098 pairs for training and 198 pairs for testing. 
%
In addition, we have collected
\NumPortraits{all} portraits from Wikiart \cite{wikiart}
(\NumPortraits{wikiart} portraits) and Wikidata \cite{wikidata}
(\NumPortraits{wikidata} portraits) without attribute annotations, and included them 
in our released benchmark, for future work on learning visual features for various genres and potentially with less human supervision.

\noindent\textbf{Baselines.} The first baseline is a simple implementation of an image-to-image style transfer model. 
The input photos are transferred into portrait style by using a UNet-based generator and a discriminator is used to play against the generator. We also compare with two state-of-the-art image generation methods, StarGAN \cite{stargan} and AttGAN \cite{AttGAN2019}.


\noindent\textbf{Experiment Setting.} 
We set the epoch number to 600 and the batch size to 32. The images are augmented by rotating, normalizing, and flipping before training to encourage the generalization of the model. The learning rate is 0.0002 and the Adam optimizer $(\beta_1=0.5, \beta_2=0.999)$ is used. The slope of leaky ReLU is 0.2.
\subsection{General Generation Quality}


\begin{table}
\centering
\begin{tabular}{lll}
\toprule
Method & IS $\uparrow$ & FID $\downarrow$ \\
\midrule
Baseline & 1.68 $\pm$ 0.148 & 0.063 \\
\emph{\Model{}} (ours) & \textbf{1.78 $\pm$ 0.182} & \textbf{0.056} \\
\bottomrule
\end{tabular}
\caption{Quantitative comparison between the baseline and proposed \emph{\Model{}} on proposed portrait dataset.}
\vspace{-20pt}
\label{tab:IS&FID}
\end{table}

\begin{table*}
\centering
\scalebox{0.78}{
\begin{tabular}{lllllllllllll}
\toprule
\parbox{0.5cm}{Evaluation Method} & Age & Clothing & Face & Gender & Hair & Mood & Pose & Setting & Style & Time & Weather & \textbf{Average} \\
\midrule
Computer & 0.88 & 0.78 & 0.82 & 0.99 & 0.86 & 0.72 & 0.59 & 0.54  & 0.95 & 0.56 & 0.83 & 0.78\\
Random & 0.20 & 0.07 & 0.07 & 0.33 & 0.17 & 0.14 & 0.10 & 0.10  & 0.13 & 0.33 & 0.10 & 0.16\\
\bottomrule
\end{tabular}}
\caption{
Attribute reconstruction accuracy by calculating the F-score of the attributes of the system generated portrait and the ground-truth attributes from the human generated portraits.
}
\vspace{-25pt}
\label{tab:Fscore}
\end{table*}


\begin{figure}[!ht]
\centering
\scalebox{0.75}{
\setlength{\tabcolsep}{1pt}
\begin{tabular}{cccccc} 
& \ctab{\includegraphics[height=1.8cm,keepaspectratio]{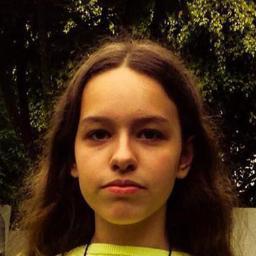}} &
\ctab{\includegraphics[height=1.8cm,keepaspectratio]{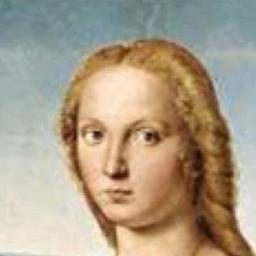}} &
\ctab{\includegraphics[height=1.8cm,keepaspectratio]{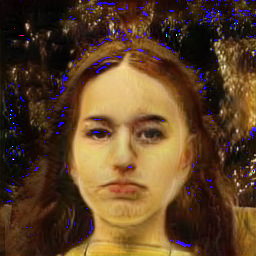}} &  &  \\
& Input Photo & \parbox[c][0.7cm]{1.8cm}{\centering Human\\ \vspace{-5pt} Generated} & Baseline & & \\
\rotatebox[origin=c]{90}{\small{\emph{\Model{}}}} &
\ctab{\includegraphics[height=1.8cm,keepaspectratio]{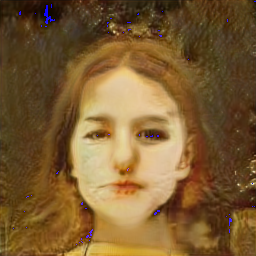}} &
\ctab{\includegraphics[height=1.8cm,keepaspectratio]{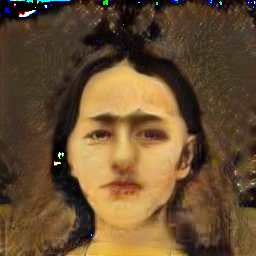}} &
\ctab{\includegraphics[height=1.8cm,keepaspectratio]{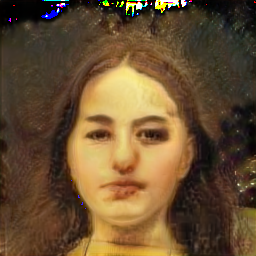}} &
\ctab{\includegraphics[height=1.8cm,keepaspectratio]{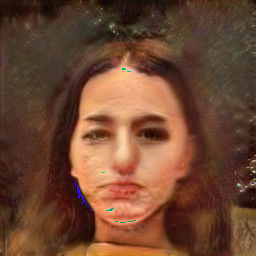}} &
\ctab{\includegraphics[height=1.8cm,keepaspectratio]{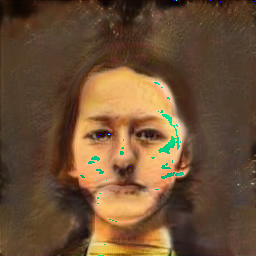}} \\
\rotatebox[origin=c]{90}{\small{StarGAN}} &
\ctab{\includegraphics[height=1.8cm,keepaspectratio]{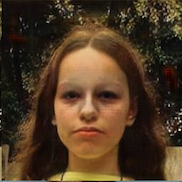}} &
\ctab{\includegraphics[height=1.8cm,keepaspectratio]{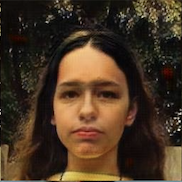}} &
\ctab{\includegraphics[height=1.8cm,keepaspectratio]{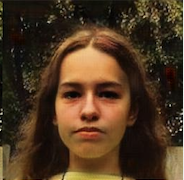}} &
\ctab{\includegraphics[height=1.8cm,keepaspectratio]{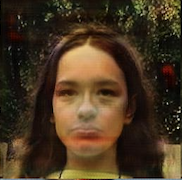}} &
\ctab{\includegraphics[height=1.8cm,keepaspectratio]{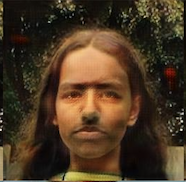}} \\
\rotatebox[origin=c]{90}{\small{AttGAN}} &
\ctab{\includegraphics[height=1.8cm,keepaspectratio]{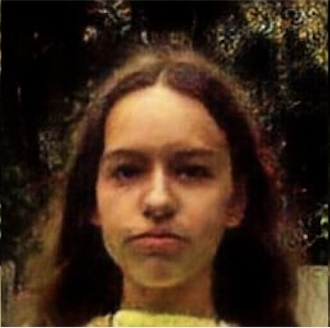}} &
\ctab{\includegraphics[height=1.8cm,keepaspectratio]{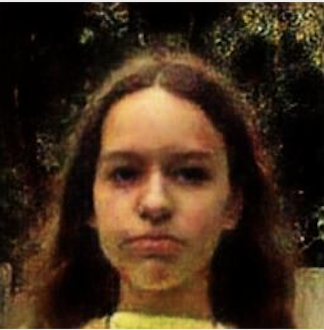}} &
\ctab{\includegraphics[height=1.8cm,keepaspectratio]{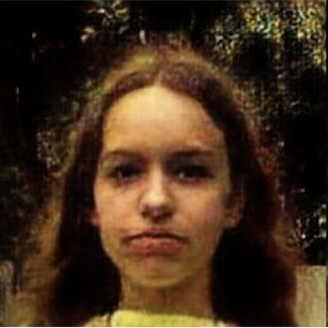}} &
\ctab{\includegraphics[height=1.8cm,keepaspectratio]{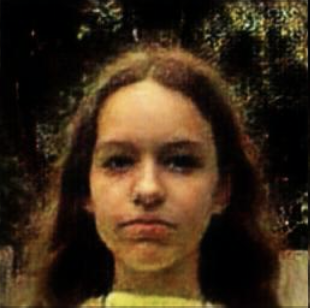}} &
\ctab{\includegraphics[height=1.8cm,keepaspectratio]{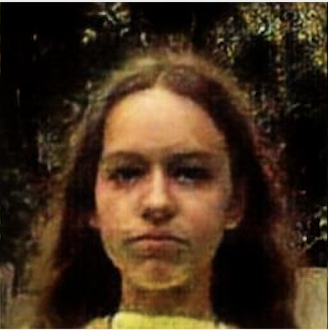}} \\
& Blond Hair & Black Hair & Wavy Hair & Straight Hair & Short Hair \\
\end{tabular}}
\caption{Synthetic portraits of the proposed \emph{\Model{}}, StarGAN \cite{stargan} and AttGAN \cite{AttGAN2019} trained on the portrait dataset given hair color.}
\vspace{-15pt}
\label{fig:hair}
\end{figure}

\begin{figure}[t]
\centering
\scalebox{0.75}{
\setlength{\tabcolsep}{1pt}
\begin{tabular}{cccccc}
& \ctab{\includegraphics[height=1.8cm,keepaspectratio]{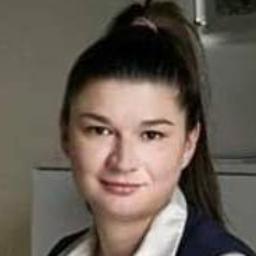}} &
\ctab{\includegraphics[height=1.8cm,keepaspectratio]{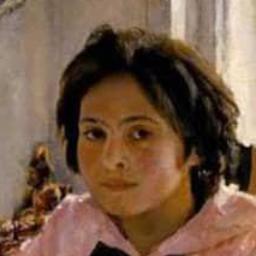}} &
\ctab{\includegraphics[height=1.8cm,keepaspectratio]{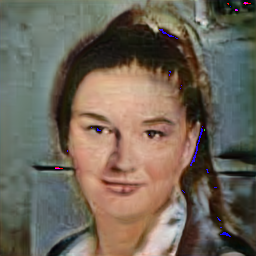}} & & \\
& Input Photo & \parbox[c][0.7cm]{1.8cm}{\centering Human\\ \vspace{-5pt} Generated} & Baseline &  & \\
\rotatebox[origin=c]{90}{\small{\emph{\Model{}}}} &
\ctab{\includegraphics[height=1.8cm,keepaspectratio]{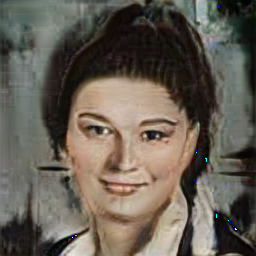}} &
\ctab{\includegraphics[height=1.8cm,keepaspectratio]{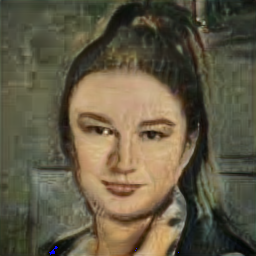}} &
\ctab{\includegraphics[height=1.8cm,keepaspectratio]{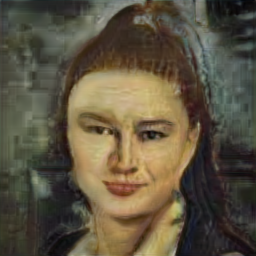}} &
\ctab{\includegraphics[height=1.8cm,keepaspectratio]{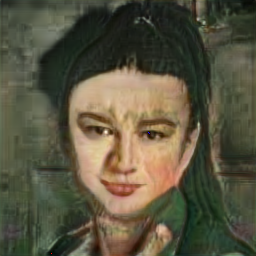}} &
\ctab{\includegraphics[height=1.8cm,keepaspectratio]{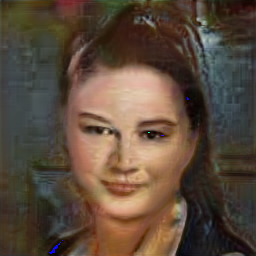}} \\
\rotatebox[origin=c]{90}{\small{StarGAN}} &
\ctab{\includegraphics[height=1.8cm,keepaspectratio]{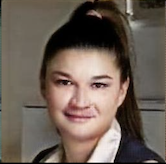}} &
\ctab{\includegraphics[height=1.8cm,keepaspectratio]{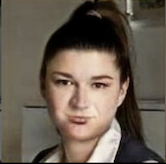}} &
\ctab{\includegraphics[height=1.8cm,keepaspectratio]{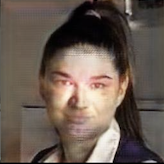}} &
\ctab{\includegraphics[height=1.8cm,keepaspectratio]{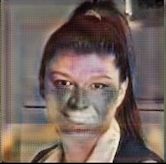}} &
\ctab{\includegraphics[height=1.8cm,keepaspectratio]{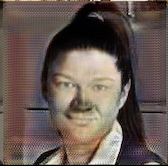}} \\
\rotatebox[origin=c]{90}{\small{AttGAN}} &
\ctab{\includegraphics[height=1.8cm,keepaspectratio]{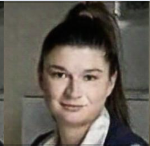}} &
\ctab{\includegraphics[height=1.8cm,keepaspectratio]{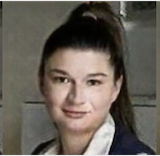}} &
\ctab{\includegraphics[height=1.8cm,keepaspectratio]{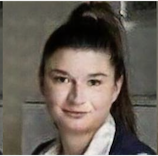}} &
\ctab{\includegraphics[height=1.8cm,keepaspectratio]{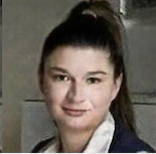}} &
\ctab{\includegraphics[height=1.8cm,keepaspectratio]{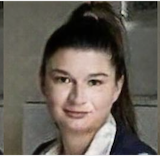}} \\
& Smile & \parbox[c][0.7cm]{1.8cm}{\centering Blank\\ \vspace{-5pt} Expression} & Glance & \parbox[c][0.7cm]{1.8cm}{\centering Quizzical\\ \vspace{-5pt} Expression} & Smirk \\
\end{tabular}}
\caption{Synthetic portraits of the proposed \emph{\Model{}}, StarGAN \cite{stargan} and AttGAN \cite{AttGAN2019} given an attribute of face expression.}
\vspace{-15pt}
\label{fig:face}
\end{figure}

\begin{figure}[t]
\centering
\scalebox{0.8}{
\setlength{\tabcolsep}{1pt}
\begin{tabular}{ccccc}
& \ctab{\includegraphics[height=1.8cm,keepaspectratio]{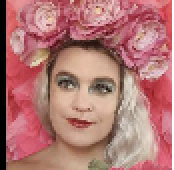}} &
\ctab{\includegraphics[height=1.8cm,keepaspectratio]{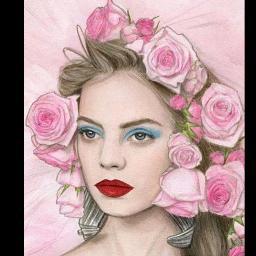}} &
\ctab{\includegraphics[height=1.8cm,keepaspectratio]{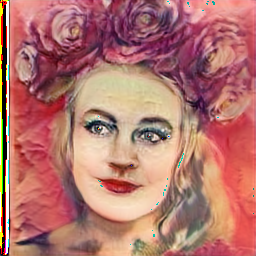}} & \\
& Input Photo & \parbox[c][0.7cm]{1.8cm}{\centering Human\\ \vspace{-5pt} Generated} & Baseline &  \\
\rotatebox[origin=c]{90}{\small{\emph{\Model{}}}} &
\ctab{\includegraphics[height=1.8cm,keepaspectratio]{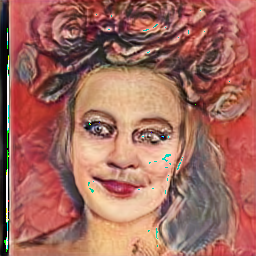}} &
\ctab{\includegraphics[height=1.8cm,keepaspectratio]{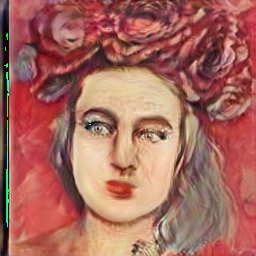}} &
\ctab{\includegraphics[height=1.8cm,keepaspectratio]{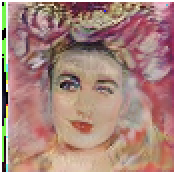}} &
\ctab{\includegraphics[height=1.8cm,keepaspectratio]{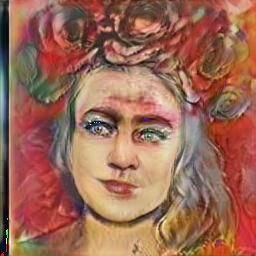}} \\
\rotatebox[origin=c]{90}{\small{StarGAN}} &
\ctab{\includegraphics[height=1.8cm,keepaspectratio]{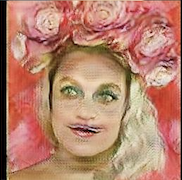}} &
\ctab{\includegraphics[height=1.8cm,keepaspectratio]{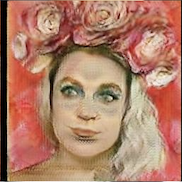}} &
\ctab{\includegraphics[height=1.8cm,keepaspectratio]{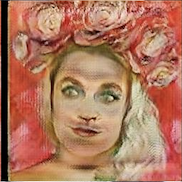}} &
\ctab{\includegraphics[height=1.8cm,keepaspectratio]{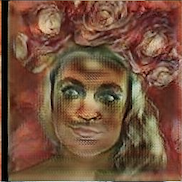}} \\
\rotatebox[origin=c]{90}{\small{AttGAN}} &
\ctab{\includegraphics[height=1.8cm,keepaspectratio]{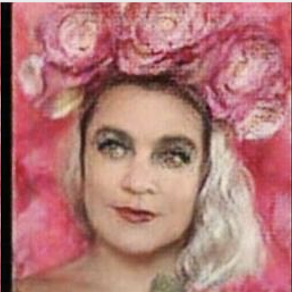}} &
\ctab{\includegraphics[height=1.8cm,keepaspectratio]{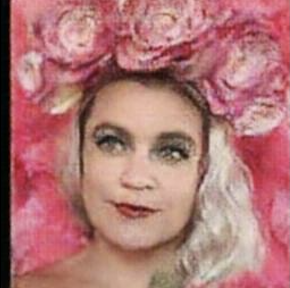}} &
\ctab{\includegraphics[height=1.8cm,keepaspectratio]{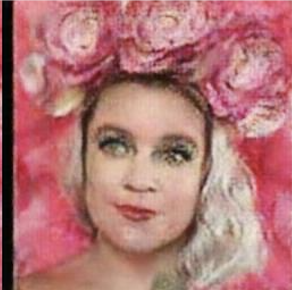}} &
\ctab{\includegraphics[height=1.8cm,keepaspectratio]{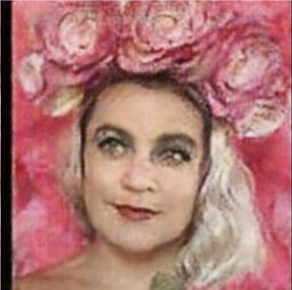}} \\
& Happy & Calm & Excited & Sad \\
\end{tabular}}
\caption{Synthetic portraits of the proposed \emph{\Model{}}, StarGAN \cite{stargan} and AttGAN \cite{AttGAN2019} given an attribute of mood expression.}
\vspace{-15pt}
\label{fig:mood}
\end{figure}

\begin{figure*}[t]
\centering
\scalebox{0.85}{
\setlength{\tabcolsep}{1pt}
\begin{tabular}{ccccccccc}
\ctab{\includegraphics[height=1.8cm,keepaspectratio]{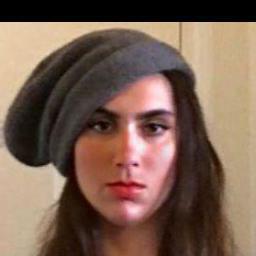}} &
\ctab{\includegraphics[height=1.8cm,keepaspectratio]{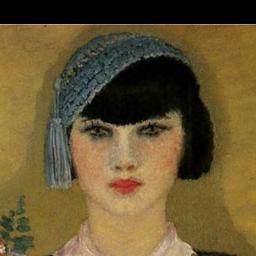}} &
\ctab{\includegraphics[height=1.8cm,keepaspectratio]{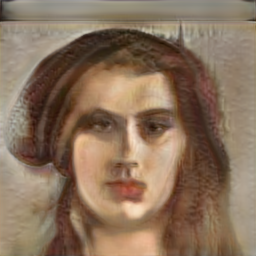}} &
\ctab{\includegraphics[height=1.8cm,keepaspectratio]{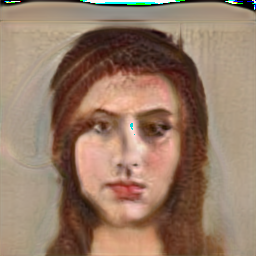}} &
\ctab{\includegraphics[height=1.8cm,keepaspectratio]{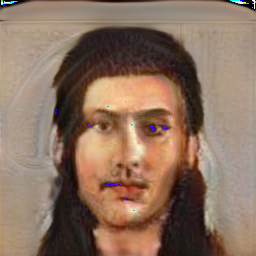}} &
\ctab{\includegraphics[height=1.8cm,keepaspectratio]{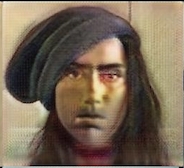}} &
\ctab{\includegraphics[height=1.8cm,keepaspectratio]{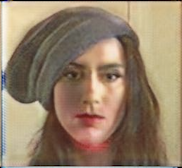}} &
\ctab{\includegraphics[height=1.8cm,keepaspectratio]{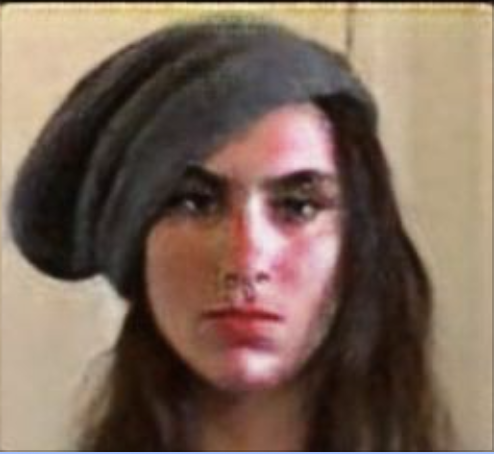}} &
\ctab{\includegraphics[height=1.8cm,keepaspectratio]{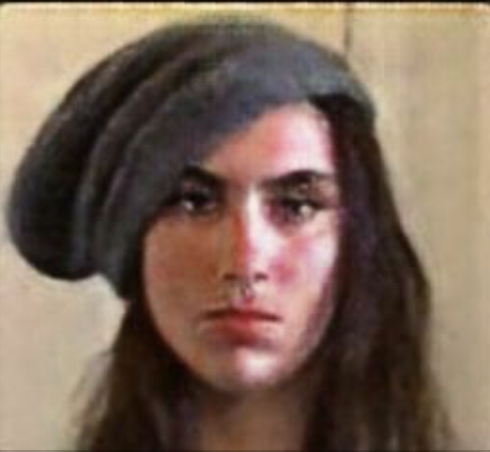}} \\
Input Photo & \parbox{1.8cm}{\centering Human Generated} & Baseline & \parbox{1.8cm}{\centering\emph{\Model{}} \\ w. Female} & \parbox{1.8cm}{\centering\emph{\Model{}} \\ w. Male} & \parbox{1.8cm}{\centering StarGAN \\ w. Female} & \parbox{1.8cm}{\centering StarGAN \\ w. Male} & \parbox{1.8cm}{\centering AttGAN \\ w. Female} & \parbox{1.8cm}{\centering AttGAN \\ w. Male} \\
\end{tabular}}
\caption{Synthetic portraits of the proposed \emph{\Model{}}, StarGAN \cite{stargan} and AttGAN \cite{AttGAN2019} given an attribute of gender. 
}
\vspace{-15pt}
\label{fig:gender}
\end{figure*}

A good portrait must attract our eyes in some way: its subject matter, its use of color, an interesting juxtaposition of objects, its realistic appearance, 
or any number of other factors. We evaluate the general visual quality of a portrait using the standard metric Inception Score (IS) \cite{Incept_score} based on ImageNet \cite{imagenet} predefined classes. Suppose $\mathcal{V}_g$ is the collection of generated portraits from the last layer of the generator $G_M$ from the generation stack. We feed all generated portraits $v\in\mathcal{V}_g$ into Inceptive-v3 networks \cite{Inception} to obtain their conditional probability distributions $p(s|v)$ over 1000 ImageNet~\cite{imagenet} classes, where each class is denoted by $s$. Class distribution is then marginalized by assuming uniform distribution of $v\in \mathcal{V}_g$, i.e.
\vspace{-10pt}
\begin{equation*}
    q(s) = \sum_{v\in\mathcal{V}_g}p(s|v)\frac{1}{|\mathcal{V}_g|}.
\vspace{-10pt}
\end{equation*}
IS score is the average KL-divergence between $p_v=p(\cdot|v)$ and marginal distribution $q(\cdot)$,

\begin{equation*}
    \mathrm{IS}(\mathcal{V}_g) = \frac{1}{|\mathcal{V}_g|}\sum_{v\in\mathcal{V}_g}D_\mathrm{KL}(p_v\|q)
\vspace{-10pt}
\end{equation*}

A higher IS indicates the generated images are more realistic in the sense that their conditional distributions concentrate on a small subset of classes. Although IS is widely used, it does not compare the generated results with real samples \cite{FID}. Fr\'echet Inception Distance (FID) \cite{FID} is another popular metric for conditioned image generation, which measures the Fr\'echet distance between the generated and real (gold standard) image distribution. Lower FID is better, indicating the generated results and target samples are more similar. 

Table \ref{tab:IS&FID} shows the proposed \emph{\Model{}} outperforms the baseline with the IS score increased by 6\% and the FID score decreased by 11\%. By taking attribute embeddings as additional input, the model can build the alignment between the input photo and target portrait more easily, making the hidden representation more meaningful to produce more reasonable composition. E.g., the attribute value \texttt{blond hair} can provide clearer guidance of generating a particular hair color rather than some unpredictable behavior. 
\subsection{Illustrating Textual Attributes}

\begin{figure}[t]
\centering
\scalebox{0.95}{
\setlength{\tabcolsep}{1pt}
\begin{tabular}{ccccc}
\rotatebox[origin=c]{90}{\small{Input Photo}} & 
\ctab{\includegraphics[height=1.8cm,keepaspectratio]{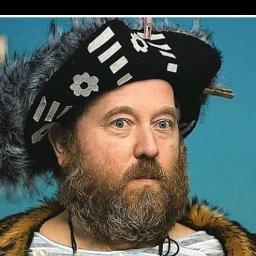}} &
\ctab{\includegraphics[height=1.8cm,keepaspectratio]{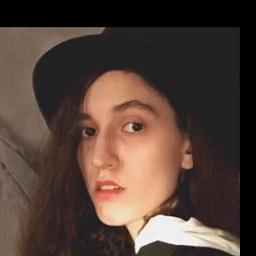}} &
\ctab{\includegraphics[height=1.8cm,keepaspectratio]{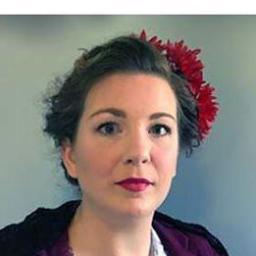}} &
\ctab{\includegraphics[height=1.8cm,keepaspectratio]{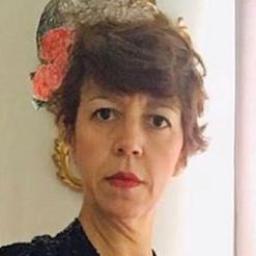}} \\
\rotatebox[origin=c]{90}{\scriptsize{Human Generated}} &
\ctab{\includegraphics[height=1.8cm,keepaspectratio]{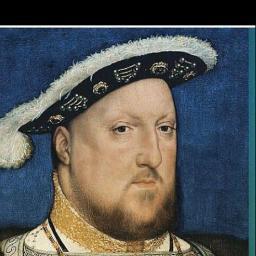}} &
\ctab{\includegraphics[height=1.8cm,keepaspectratio]{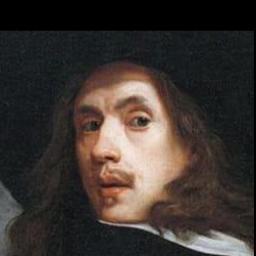}} &
\ctab{\includegraphics[height=1.8cm,keepaspectratio]{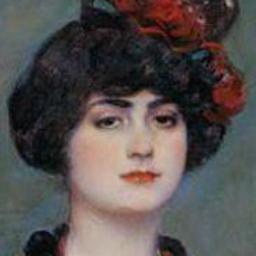}} &
\ctab{\includegraphics[height=1.8cm,keepaspectratio]{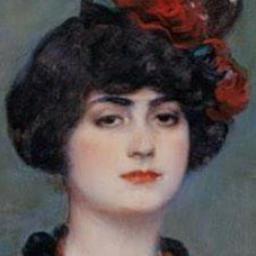}} \\
\rotatebox[origin=c]{90}{\small{Baseline}} & 
\ctab{\includegraphics[height=1.8cm,keepaspectratio]{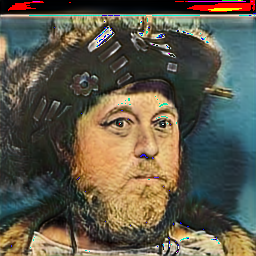}} &
\ctab{\includegraphics[height=1.8cm,keepaspectratio]{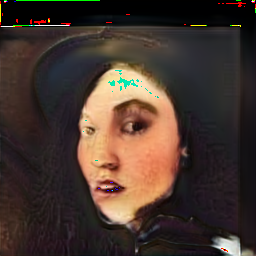}} &
\ctab{\includegraphics[height=1.8cm,keepaspectratio]{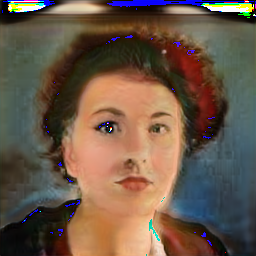}} &
\ctab{\includegraphics[height=1.8cm,keepaspectratio]{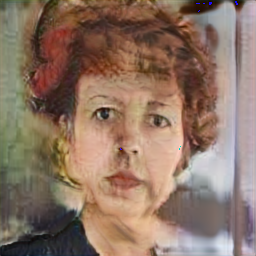}} \\
\rotatebox[origin=c]{90}{\scriptsize{System Generated}} &
\ctab{\includegraphics[height=1.8cm,keepaspectratio]{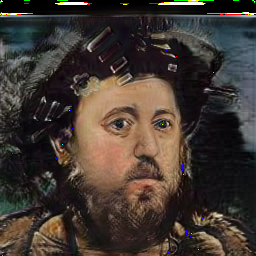}} &
\ctab{\includegraphics[height=1.8cm,keepaspectratio]{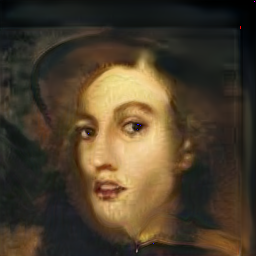}} &
\ctab{\includegraphics[height=1.8cm,keepaspectratio]{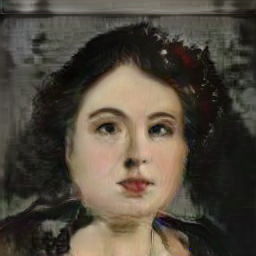}} &
\ctab{\includegraphics[height=1.8cm,keepaspectratio]{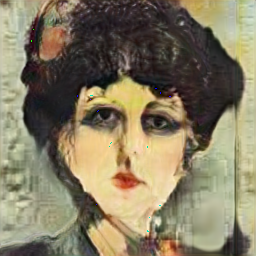}} \\
& 
\scriptsize{\parbox{1.8cm}{\centering Calm; Classical Art; Sitting; Rainy; Cold; Male; Black hair; Short hair}} &
\scriptsize{\parbox{1.8cm}{\centering Young adults; Calm; Classical Art; Before 1970; Stormy; Frown; Glance; Male; Coat; Blond hair}} &
\scriptsize{\parbox{1.8cm}{\centering Calm; Classical Art; Sitting; Cold; Blank Expression; Female; Dress; Wavy hair; Black hair}} & 
\scriptsize{\parbox{1.8cm}{\centering Calm; Impressionism; Sitting; Stormy; Cloudy; Female; Dress; Black hair}}
\\
\end{tabular}}
\caption{Synthetic portraits of the proposed \emph{\Model{}} given combined attributes including age, mood, style, time, pose, setting, weather, face expression, gender, clothing and hair.}
\label{fig:combined}
\end{figure}

\begin{figure}[t]
\centering
\scalebox{0.75}{
\setlength{\tabcolsep}{1pt}
\begin{tabular}{cccccc}
\rotatebox[origin=c]{90}{\small{Input Photo}} &
\ctab{\includegraphics[height=1.8cm,keepaspectratio]{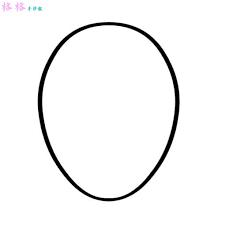}} &
\ctab{\includegraphics[height=1.8cm,keepaspectratio]{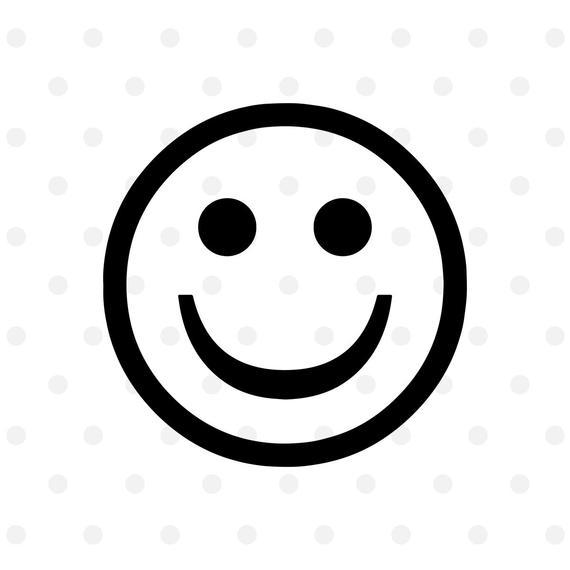}} &
\ctab{\includegraphics[height=1.8cm,keepaspectratio]{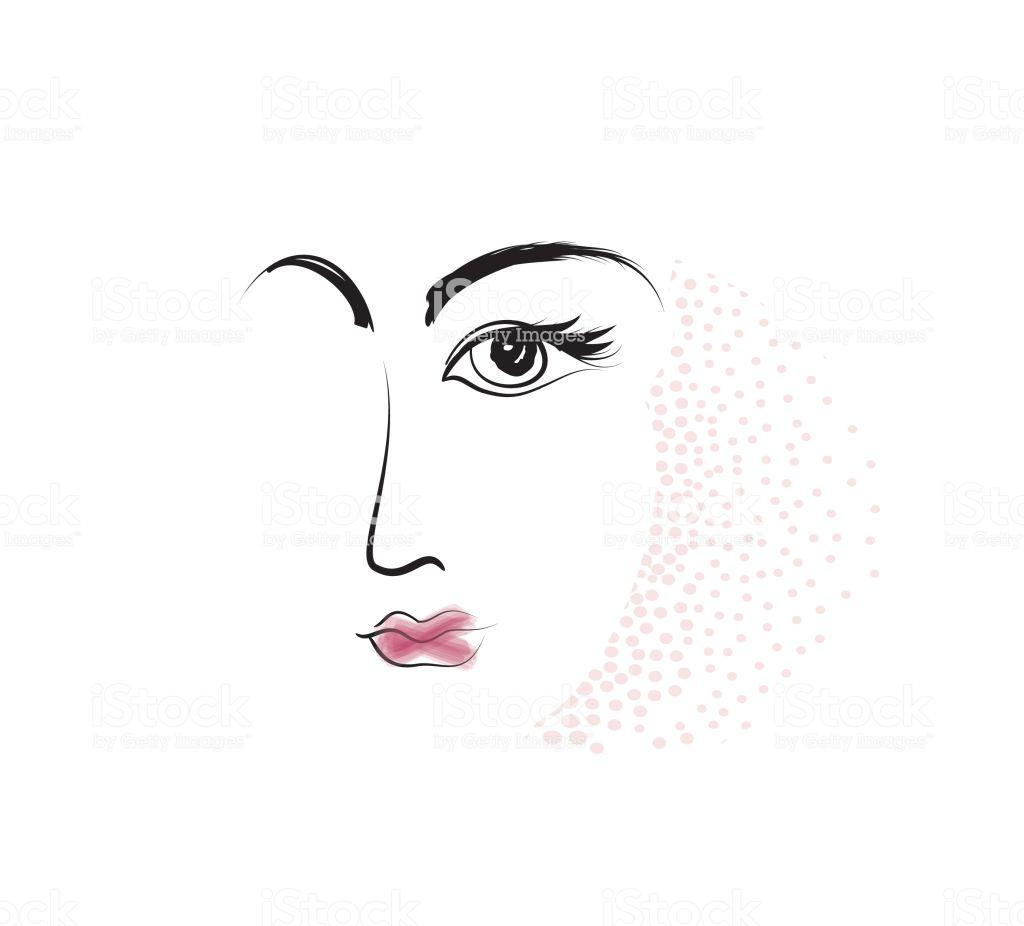}} &
\ctab{\includegraphics[height=1.8cm,keepaspectratio]{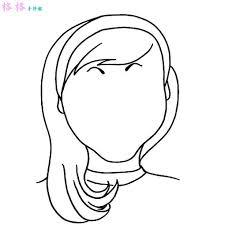}} &
\ctab{\includegraphics[height=1.8cm,keepaspectratio]{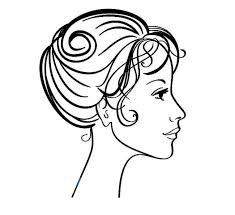}} \\
\rotatebox[origin=c]{90}{\small{Blond Hair}} &
\ctab{\includegraphics[height=1.8cm,keepaspectratio]{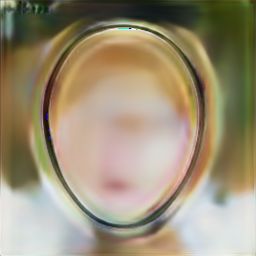}} &
\ctab{\includegraphics[height=1.8cm,keepaspectratio]{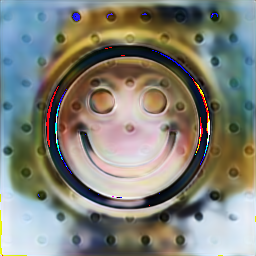}} &
\ctab{\includegraphics[height=1.8cm,keepaspectratio]{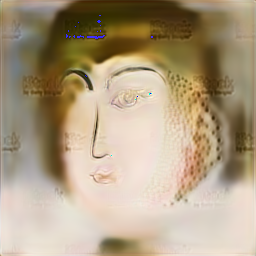}} &
\ctab{\includegraphics[height=1.8cm,keepaspectratio]{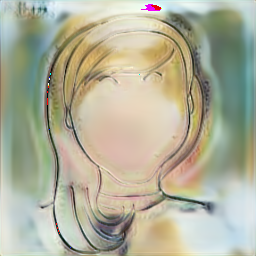}} &
\ctab{\includegraphics[height=1.8cm,keepaspectratio]{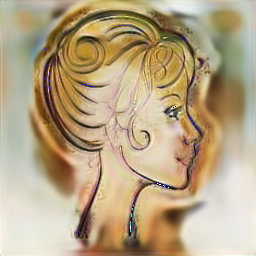}} \\
\rotatebox[origin=c]{90}{\small{Black Hair}} &
\ctab{\includegraphics[height=1.8cm,keepaspectratio]{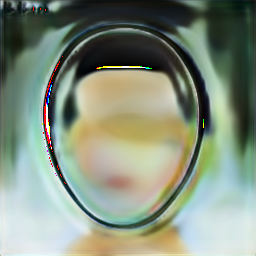}} &
\ctab{\includegraphics[height=1.8cm,keepaspectratio]{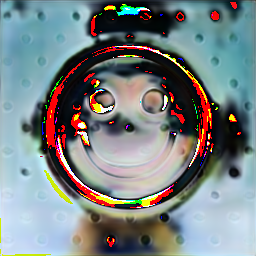}} &
\ctab{\includegraphics[height=1.8cm,keepaspectratio]{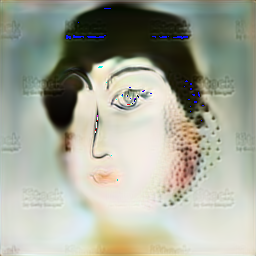}} &
\ctab{\includegraphics[height=1.8cm,keepaspectratio]{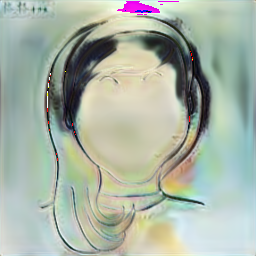}} &
\ctab{\includegraphics[height=1.8cm,keepaspectratio]{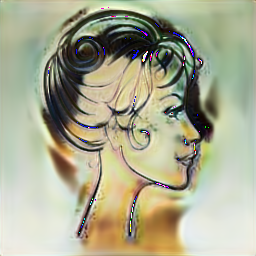}} \\
\end{tabular}}
\caption{Given drawn sketches and an attribute of hair color or weather, the attribute value can be correctly grounded in the generated portraits.}
\vspace{-15pt}
\label{fig:grounding}
\end{figure}

\begin{figure}[t]
\centering
\scalebox{0.75}{
\setlength{\tabcolsep}{1pt}
\begin{tabular}{ccccc} 
\ctab{\includegraphics[height=1.8cm,keepaspectratio]{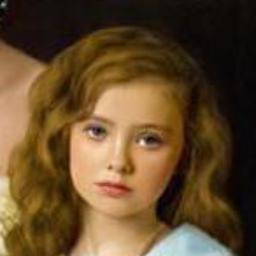}} &
\ctab{\includegraphics[height=1.8cm,keepaspectratio]{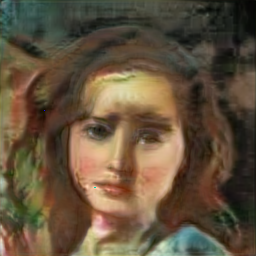}} &
\ctab{\includegraphics[height=1.8cm,keepaspectratio]{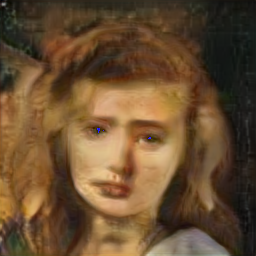}} &
\ctab{\includegraphics[height=1.8cm,keepaspectratio]{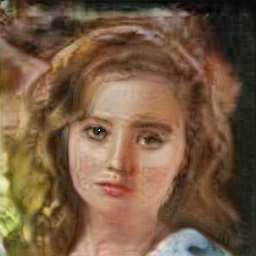}} &
\ctab{\includegraphics[height=1.8cm,keepaspectratio]{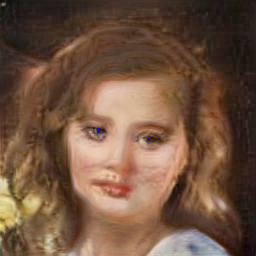}} \\
\ctab{\includegraphics[height=1.8cm,keepaspectratio]{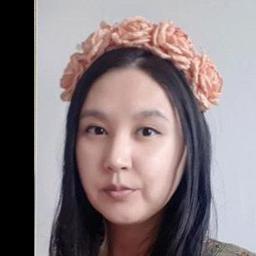}} &
\ctab{\includegraphics[height=1.8cm,keepaspectratio]{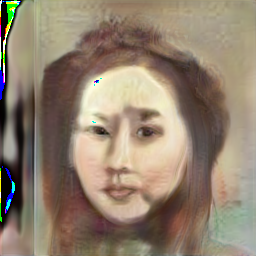}} &
\ctab{\includegraphics[height=1.8cm,keepaspectratio]{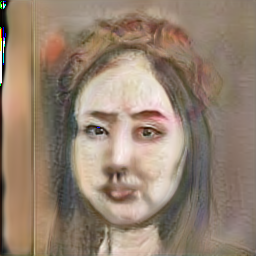}} &
\ctab{\includegraphics[height=1.8cm,keepaspectratio]{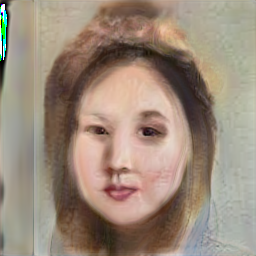}} &
\ctab{\includegraphics[height=1.8cm,keepaspectratio]{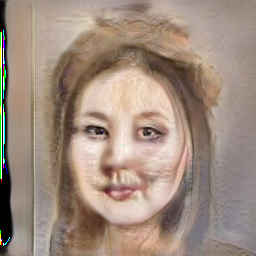}} \\
\ctab{\includegraphics[height=1.8cm,keepaspectratio]{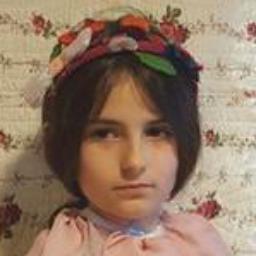}} &
\ctab{\includegraphics[height=1.8cm,keepaspectratio]{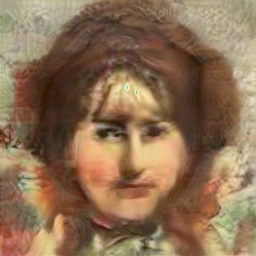}} &
\ctab{\includegraphics[height=1.8cm,keepaspectratio]{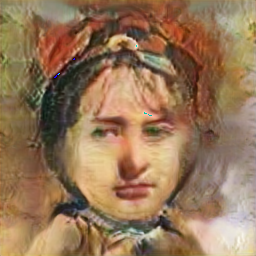}} &
\ctab{\includegraphics[height=1.8cm,keepaspectratio]{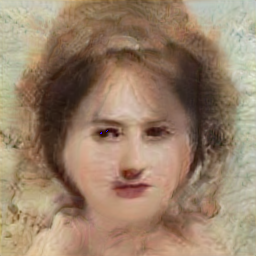}} &
\ctab{\includegraphics[height=1.8cm,keepaspectratio]{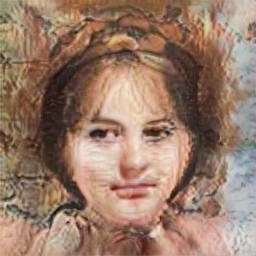}} \\
\ctab{\includegraphics[height=1.8cm,keepaspectratio]{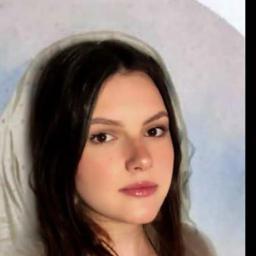}} &
\ctab{\includegraphics[height=1.8cm,keepaspectratio]{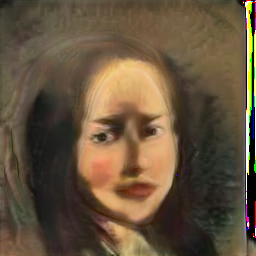}} &
\ctab{\includegraphics[height=1.8cm,keepaspectratio]{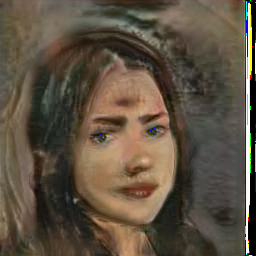}} &
\ctab{\includegraphics[height=1.8cm,keepaspectratio]{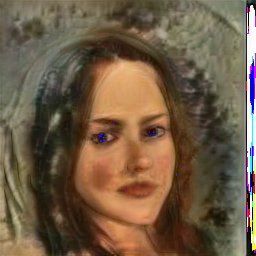}} &
\ctab{\includegraphics[height=1.8cm,keepaspectratio]{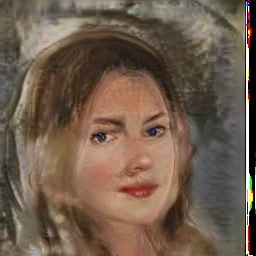}} \\
Input Photo & Stormy & Sad & Sunny & Happy \\
\end{tabular}}
\caption{Inter-dependency of attributes.}
\vspace{-15pt}
\label{fig:dependency}
\end{figure}





We propose a new metric to check if the generated portraits reflect the input attribute values, and further explore how attribute semantics are learned and grounded into generated portraits.


\noindent\textbf{Subject Attribute Reconstruction Accuracy.} 
Since most of ImageNet classes are objects and the number of classes that appear in portraits is quite limited, the IS metric cannot evaluate the subtle details of human subjects in portraits. 
We use our attribute classifier $F$ trained on the portrait data set to classify each generated portrait. We compute the F-score of the estimated attributes against the textual attributes extracted from the human generated portrait. Table \ref{tab:Fscore} shows the attribute reconstruction accuracy.
We can see that our approach successfully illustrates 78\% attributes.


\noindent\textbf{Single Attribute Coherence} We first consider a simpler scenario where we change only one attribute. We train a separate model for each attribute, for which we constrain the attribute embeddings $\mathbf{v}$ to contain only the specific attribute (e.g. \texttt{hair}) during training. 
\Cref{fig:hair,fig:face,fig:mood,fig:gender}
show example outputs of the proposed \emph{\Model{}} compared with StarGAN \cite{stargan} and AttGAN \cite{AttGAN2019} by modifying \texttt{hair}, \texttt{face expression}, \texttt{mood} and \texttt{gender} 
respectively. 
We can see our models have successfully changed 
these attributes and 
learned to generate according to attribute values rather than performing some random behavior. In contrast, the StarGAN mistakenly interprets the attributes and AttGAN does not learn from the attributes at all, due to the small amount of training data and lacking of the mechanisms for cross-domain (photo to portrait) image transformation.

\noindent\textbf{Multiple Attributes} Here we consider a more complex scenario that we change all attributes for portrait generation. Figure \ref{fig:combined} shows examples using a combination of 11 attributes listed in Table \ref{tab:dimensions}. 
\emph{\Model{}} can capture the explicit attribute values such as \texttt{blond} and \texttt{black} hair, and 
capture the abstract concepts such as \texttt{stormy} vs \texttt{sunny} and \texttt{cold} vs \texttt{hot} by adjusting the background darkness levels.

\noindent\textbf{Grounding of Attributes}
We further examine how the model ground attributes into generated portraits, by taking an demonstrative example of the \texttt{hair} attribute as shown in Figure \ref{fig:grounding}. 
From these results we observe that \emph{\Model{}} can rely on either face outlines or facial features such as eyes or mouths to estimate the relative position of components including hair and facial area. In the fourth column we design a blank face with hair outline. Although \emph{\Model{}} cannot perfectly color the hair area, we do observe the model tends to color along the lines. We also show in the last column that \emph{\Model{}} works in a similar way for faces in profile.

\noindent\textbf{Inter-dependency of Attributes}
As discussed in section~\ref{sec:EntityTyping}, the attribute values in Table \ref{tab:dimensions} are correlated. Here we show some examples of two related attributes, \texttt{weather} and \texttt{mood}, in Figure \ref{fig:dependency}. For example, 
the \texttt{happy} portrait in the same row is associated with a brighter background implying a sunny weather. 
We also use the pre-trained classifier $F$ to quantitatively evaluate the inter-dependency between \texttt{weather} and \texttt{mood}. Given the \texttt{happy} portraits, the predicted probabilities to be \texttt{sunny} and \texttt{stormy} are 38.28\% and 0.22\%, respectively. Similarly, given the \texttt{sunny} portraits, the predicted probabilities to be \texttt{happy} and \texttt{sad} are 39.20\% and 0.31\%, respectively.


\noindent\textbf{Affordances of Attributes}
Not all the compositions of attributes are semantically valid. For example, it is not possible to have ``smile'' face expression and ``sad'' mood in the same portrait. Although it is challenging to inform a model with such affordances, we can use visual signal to improve the generalization capabilities of the model since invalid combinations will not appear in the visual domain. Table \ref{tab:affordances} shows the attribute reconstruction accuracy, IS score and the FID score of the generated ``smile sad'' and ``smile happy'' portraits. Using a valid attributes combination, the model can generate a portrait with better quality and better reconstruction accuracy.

\begin{table}
\centering
\footnotesize
\begin{tabular}{lllll}
\toprule
\multirow{2}{*}{Attributes} & \multicolumn{2}{c}{Recons. Acc. $\uparrow$} & \multirow{2}{*}{IS $\uparrow$} & \multirow{2}{*}{FID $\downarrow$} \\
 & Mood & Face &   &   \\
\midrule
Smile+Sad & 0.960 & 0.934 & 1.76 $\pm$ 0.193 & 0.085 \\
Smile+Happy & \textbf{1.000} & \textbf{0.969} & \textbf{1.84 $\pm$ 0.212} & \textbf{0.073} \\
\bottomrule
\end{tabular}
\caption{Attributes affordances of the proposed \emph{\Model{}} evaluated on various combination of mood and face expression.}
\vspace{-25pt}
\label{tab:affordances}
\end{table}
\subsection{Remaining Challenges}
Good art works should reflect not only explicit information of the subject but also implicit attributes such as personality, occasion, occupation, nationality, hobby. However, the portrait attribute types in the proposed data set are limited due to the lack of knowledge of the subject and the artist. Automatically extracting the implicit information from professionally written text descriptions of the portraits can enrich the attribute types and further enrich the generated portraits. In addition, instead of generating only face regions, artists often include 
more complete portrait paintings containing background landscape, pose \& gestures and other objects. However, with the small amount of training samples, it is difficult to simultaneously handle all styles and content with large variation.

\section{Related Work}



Generative Adversarial Networks (GAN) \cite{NIPSGAN} achieve great success in generating realistic images 
without much control \cite{MirzaO14}.
Image-to-image generation~\cite{Isola2017pix2pix,Zhu2017CycleGAN} takes an image as input condition, and 
transfers it into another image in a different domain. 
Another line of work takes natural language as input condition and generates images accordingly. \cite{ReedAYLSL16} first introduces a conditional DC-GAN architecture 
which achieves positive results for generating low-resolution images ($64 \times 64$), but it is not equally successful in higher resolution image generation. To address this problem, Zhang \etal \cite{ZhangXLZHWM16} propose Stacked Generative Adversarial Networks (StackGAN) that first generate low resolution images and make refinements thereafter. \cite{zhang2018photographic} further proposes hierarchically-nested losses that refines images in multiple steps. \cite{Lao2019DualAI} uses dual inference over conditional and unconditional latent variables for disentanglement of content and style. Rather than starting from low-resolution image generation, \cite{hong2018inferring} first constructs semantic layouts from text and generates images based on them. Despite promising results, the above methods only incorporate sentence-level features without considering fine-grained attributes and thus yield unsatisfactory results when the input sentences are complex. 

Attentional GAN (AttnGAN)~\cite{AttnGAN} enables the generative networks to be trained on words of higher relevance, and develops an \textit{inter-modal} attention mechanism to compute the similarity between the generated image and the relevant text description. 
Recent methods~\cite{qiao2019mirrorgan,tan2019semantics} incorporate attention mechanism to improve semantic consistency. 
They rely on general caption-type instructions to generate images of flowers~\cite{NilsbackZ08}, birds~\cite{wah2011caltech} or common objects~\cite{LinMBHPRDZ14}. However, such instructions usually lack of identity information, which makes these models impossible to generate images of a specific person, flower or bird. 
In contrast, we propose to take both photo and textual attributes as input. 
It is worth noting that although AttGAN 
also leverages facial attributes when generating images, our model is different from the following aspects: (1) our goal is to generate creative and abstract portraits instead of generating photo-realistic images. (2) we apply more 
complex and abstract attributes to guide the portrait generation. We include more values for each attribute type and some of them have shared semantics; our attributes are carefully designed for portraits and are more abstract (e.g., mood); 
we use trainable attribute embeddings to better represent the inter-dependency between multiple attribute values.

\cite{Zhang2019} describes an interesting user study to evaluate the results of generating animations from screenplays where users are asked to evaluate, on a five-point Likert scale~\cite{Likert1932}, if the video shown was a reasonable animation for the text, how much of the text information was depicted in the video and how much of the information in the video was present in the text. Our attribute reconstruction metric aims at a similar goal, but we compute the scores based on pre-defined attribute categories and thus our metric is more objective.




\section{Conclusions and Future Work}

We have developed a novel method, \emph{\Model{}}, which can generate portrait paintings guided by textual attributes. 
In the future we plan to extend \emph{\Model{}} to unstructured text descriptions and apply open-domain attribute extraction techniques to extract as input, and extend it to cover a wider range of entity types and attribute types. 






\section*{Acknowledgment}
This research is based upon work supported by U.S. DARPA GAILA Program HR00111990058. The views and conclusions contained herein are those of the authors and should not be interpreted as necessarily representing the official policies, either expressed or implied, of DARPA, or the U.S. Government. The U.S. Government is authorized to reproduce and distribute reprints for governmental purposes notwithstanding any copyright annotation therein.



{\small
\bibliographystyle{unsrt} 
\bibliography{aiart}

\begin{thebibliography}{10}

\bibitem{Liao2017ACM}
Jing Liao, Yuan Yao, Lu~Yuan, Gang Hua, and Sing~Bing Kang.
\newblock Visual attribute transfer through deep image analogy.
\newblock {\em ACM Transactions on Graphics}, 36(4), 2017.

\bibitem{Zhu2017CycleGAN}
Jun{-}Yan Zhu, Taesung Park, Phillip Isola, and Alexei~A. Efros.
\newblock Unpaired image-to-image translation using cycle-consistent
  adversarial networks.
\newblock In {\em Proceedings of the IEEE International Conference on Computer
  Vision}, 2017.

\bibitem{Zhu2017BicycleGAN}
Jun-Yan Zhu, Richard Zhang, Deepak Pathak, Trevor Darrell, Alexei~A Efros,
  Oliver Wang, and Eli Shechtman.
\newblock Toward multimodal image-to-image translation.
\newblock In {\em Proceedings of the 2017 Advances in Neural Information
  Processing Systems}. 2017.

\bibitem{Lee2018DRIT}
Hsin{-}Ying Lee, Hung{-}Yu Tseng, Jia{-}Bin Huang, Maneesh~Kumar Singh, and
  Ming{-}Hsuan Yang.
\newblock Diverse image-to-image translation via disentangled representations.
\newblock In {\em Proceedings of the 15th European Conference on Computer
  Vision}, September 2018.

\bibitem{Ancient2019}
T.~{Qiao}, W.~{Zhang}, M.~{Zhang}, Z.~{Ma}, and D.~{Xu}.
\newblock Ancient painting to natural image: A new solution for painting
  processing.
\newblock In {\em 2019 IEEE Winter Conference on Applications of Computer
  Vision (WACV)}, pages 521--530, 2019.

\bibitem{Aymar1967}
Gordon~C. Aymar.
\newblock The art of portrait painting.
\newblock {\em Chilton Book Co., Philadelphia, p. 119}, 1967.

\bibitem{UNet2015}
Olaf Ronneberger, Philipp Fischer, and Thomas Brox.
\newblock U-net: Convolutional networks for biomedical image segmentation.
\newblock In Nassir Navab, Joachim Hornegger, William~M. Wells, and
  Alejandro~F. Frangi, editors, {\em Medical Image Computing and
  Computer-Assisted Intervention -- MICCAI 2015}, pages 234--241, Cham, 2015.
  Springer International Publishing.

\bibitem{AttGAN2019}
Zhenliang He, Wangmeng Zuo, Meina Kan, Shiguang Shan, and Xilin Chen.
\newblock {AttGAN}: Facial attribute editing by only changing what you want.
\newblock {\em IEEE Transactions on Image Processing}, 28(11):5464--5478, 2019.

\bibitem{UNet2019}
Xiaodan Hu, Mohamed~A. Naiel, Alexander Wong, Mark Lamm, and Paul Fieguth.
\newblock Runet: A robust unet architecture for image super-resolution.
\newblock In {\em 2019 IEEE/CVF Conference on Computer Vision and Pattern
  Recognition Workshops (CVPRW)}, pages 505--507, 2019.

\bibitem{Incept_score}
Tim Salimans, Ian~J. Goodfellow, Wojciech Zaremba, Vicki Cheung, Alec Radford,
  and Xi~Chen.
\newblock Improved techniques for training {GAN}s.
\newblock {\em CoRR}, abs/1606.03498, 2016.

\bibitem{mikolov2013efficient}
Tomas Mikolov, Kai Chen, Greg Corrado, and Jeff Dean.
\newblock Efficient estimation of word representations in vector space.
\newblock In Yoshua Bengio and Yann LeCun, editors, {\em 1st International
  Conference on Learning Representations, {ICLR} 2013, Scottsdale, Arizona,
  USA, May 2-4, 2013, Workshop Track Proceedings}, 2013.

\bibitem{relu2010}
Vinod Nair and Geoffrey~E. Hinton.
\newblock Rectified linear units improve restricted boltzmann machines.
\newblock In {\em Proceedings of the 27th International Conference on
  International Conference on Machine Learning}, 2010.

\bibitem{NIPSGAN}
Ian Goodfellow, Jean Pouget-Abadie, Mehdi Mirza, Bing Xu, David Warde-Farley,
  Sherjil Ozair, Aaron Courville, and Yoshua Bengio.
\newblock Generative adversarial nets.
\newblock In {\em Proceedings of the 28th Annual Conference on Neural
  Information Processing Systems}. 2014.

\bibitem{Getty}
{Getty Museum}.
\newblock \url{https://twitter.com/GettyMuseum}.
\newblock [Online].

\bibitem{mettwinning}
{Mett Winning}.
\newblock \url{https://www.instagram.com/explore/tags/mettwinning/}.
\newblock [Online].

\bibitem{pinchuk}
{Pinchuk Art Centre}.
\newblock \url{http://new.pinchukartcentre.org/}.
\newblock [Online].

\bibitem{rijksmuseum}
{Rijksmuseum}.
\newblock \url{https://www.rijksmuseum.nl/en}.
\newblock [Online].

\bibitem{TKQ}
{Tussen Kunst \& Quarantaine}.
\newblock \url{https://www.instagram.com/tussenkunstenquarantaine/}.
\newblock [Online].

\bibitem{booooooom}
{Booooooom}.
\newblock
  \url{https://www.booooooom.com/2011/09/27/remake-a-project-by-booooooom-and-adobe/}.
\newblock [Online].

\bibitem{pinterest}
{Pinterest}.
\newblock \url{https://www.pinterest.com/}.
\newblock [Online].

\bibitem{wikiart}
Wikiart.
\newblock \url{https://www.wikiart.org/}.
\newblock [Online].

\bibitem{wikidata}
Wikidata.
\newblock \url{https://www.wikidata.org/}.
\newblock [Online].

\bibitem{stargan}
Yunjey Choi, Min{-}Je Choi, Munyoung Kim, Jung{-}Woo Ha, Sunghun Kim, and
  Jaegul Choo.
\newblock Stargan: Unified generative adversarial networks for multi-domain
  image-to-image translation.
\newblock {\em CoRR}, 2017.

\bibitem{imagenet}
J.~{Deng}, W.~{Dong}, R.~{Socher}, L.~{Li}, {Kai Li}, and {Li Fei-Fei}.
\newblock {ImageNet: A Large-Scale Hierarchical Image Database}.
\newblock In {\em Proceedings of the IEEE International Conference on Computer
  Vision}, 2009.

\bibitem{Inception}
Christian Szegedy, Vincent Vanhoucke, Sergey Ioffe, Jonathon Shlens, and
  Zbigniew Wojna.
\newblock Rethinking the inception architecture for computer vision.
\newblock {\em CoRR}, abs/1512.00567, 2015.

\bibitem{FID}
Martin Heusel, Hubert Ramsauer, Thomas Unterthiner, Bernhard Nessler,
  G{\"{u}}nter Klambauer, and Sepp Hochreiter.
\newblock {GANs} trained by a two time-scale update rule converge to a nash
  equilibrium.
\newblock {\em CoRR}, abs/1706.08500, 2017.

\bibitem{MirzaO14}
Mehdi Mirza and Simon Osindero.
\newblock Conditional generative adversarial nets.
\newblock {\em CoRR}, abs/1411.1784, 2014.

\bibitem{Isola2017pix2pix}
Phillip Isola, Jun{-}Yan Zhu, Tinghui Zhou, and Alexei~A. Efros.
\newblock Image-to-image translation with conditional adversarial networks.
\newblock In {\em Proceedings of the IEEE Computer Society Conference on
  Computer Vision and Pattern Recognition}, 2017.

\bibitem{ReedAYLSL16}
Scott~E. Reed, Zeynep Akata, Xinchen Yan, Lajanugen Logeswaran, Bernt Schiele,
  and Honglak Lee.
\newblock Generative adversarial text to image synthesis.
\newblock In {\em Proceedings of the 33rd International Conference on
  International Conference on Machine Learning}, 2016.

\bibitem{ZhangXLZHWM16}
Han Zhang, Tao Xu, Hongsheng Li, Shaoting Zhang, Xiaolei Huang, Xiaogang Wang,
  and Dimitris~N. Metaxas.
\newblock Stack{GAN}: Text to photo-realistic image synthesis with stacked
  generative adversarial networks.
\newblock In {\em Proceedings of the IEEE International Conference on Computer
  Vision}, 2017.

\bibitem{zhang2018photographic}
Zizhao Zhang, Yuanpu Xie, and Lin Yang.
\newblock Photographic text-to-image synthesis with a hierarchically-nested
  adversarial network.
\newblock In {\em Proceedings of the IEEE Computer Society Conference on
  Computer Vision and Pattern Recognition}, pages 6199--6208, 2018.

\bibitem{Lao2019DualAI}
Qicheng Lao, Mohammad Havaei, Ahmad Pesaranghader, Francis Dutil, Lisa
  Di{-}Jorio, and Thomas Fevens.
\newblock Dual adversarial inference for text-to-image synthesis.
\newblock In {\em 2019 {IEEE/CVF} International Conference on Computer Vision,
  {ICCV} 2019, Seoul, Korea (South), October 27 - November 2, 2019}, pages
  7566--7575. {IEEE}, 2019.

\bibitem{hong2018inferring}
Seunghoon Hong, Dingdong Yang, Jongwook Choi, and Honglak Lee.
\newblock Inferring semantic layout for hierarchical text-to-image synthesis.
\newblock In {\em Proceedings of the IEEE Computer Society Conference on
  Computer Vision and Pattern Recognition}, pages 7986--7994, 2018.

\bibitem{AttnGAN}
Tao Xu, Pengchuan Zhang, Qiuyuan Huang, Han Zhang, Zhe Gan, Xiaolei Huang, and
  Xiaodong He.
\newblock Attn{GAN}: Fine-grained text to image generation with attentional
  generative adversarial networks.
\newblock {\em CoRR}, abs/1711.10485, 2017.

\bibitem{qiao2019mirrorgan}
Tingting Qiao, Jing Zhang, Duanqing Xu, and Dacheng Tao.
\newblock Mirrorgan: Learning text-to-image generation by redescription.
\newblock In {\em Proceedings of the IEEE Computer Society Conference on
  Computer Vision and Pattern Recognition}, 2019.

\bibitem{tan2019semantics}
Hongchen Tan, Xiuping Liu, Xin Li, Yi~Zhang, and Baocai Yin.
\newblock Semantics-enhanced adversarial nets for text-to-image synthesis.
\newblock In {\em Proceedings of the IEEE International Conference on Computer
  Vision}, pages 10500--10509, 2019.

\bibitem{NilsbackZ08}
Maria{-}Elena Nilsback and Andrew Zisserman.
\newblock Automated flower classification over a large number of classes.
\newblock In {\em Sixth Indian Conference on Computer Vision, Graphics {\&}
  Image Processing, {ICVGIP} 2008, Bhubaneswar, India, 16-19 December 2008},
  pages 722--729. {IEEE} Computer Society, 2008.

\bibitem{wah2011caltech}
Catherine Wah, Steve Branson, Peter Welinder, Pietro Perona, and Serge
  Belongie.
\newblock The caltech-ucsd birds-200-2011 dataset.
\newblock 2011.

\bibitem{LinMBHPRDZ14}
Tsung{-}Yi Lin, Michael Maire, Serge~J. Belongie, James Hays, Pietro Perona,
  Deva Ramanan, Piotr Doll{\'{a}}r, and C.~Lawrence Zitnick.
\newblock Microsoft {COCO:} common objects in context.
\newblock In David~J. Fleet, Tom{\'{a}}s Pajdla, Bernt Schiele, and Tinne
  Tuytelaars, editors, {\em Computer Vision - {ECCV} 2014 - 13th European
  Conference, Zurich, Switzerland, September 6-12, 2014, Proceedings, Part
  {V}}, volume 8693 of {\em Lecture Notes in Computer Science}, pages 740--755.
  Springer, 2014.

\bibitem{Zhang2019}
Yeyao Zhang, Eleftheria Tsipidi, Sasha Schriber, Mubbasir Kapadia, Markus
  Gross, and Ashutosh Modi.
\newblock Generating animations from screenplays.
\newblock In {\em Proceedings of the Eighth Joint Conference on Lexical and
  Computational Semantics}, 2019.

\bibitem{Likert1932}
Rensis Likert.
\newblock A technique for the measurement of attitudes.
\newblock {\em Archives of Psychology, 22(140):1–55}, 1932.

\end{thebibliography}
}

\end{document}